\documentclass{article}
\usepackage{spconf,amsmath,graphicx}
\usepackage{array}
\newcolumntype{P}[1]{>{\centering\arraybackslash}p{#1}}

\usepackage{enumitem}
\setlist{nosep, leftmargin=14pt}

\usepackage{mwe} 

\usepackage{afterpage}

\newcommand\blankpage{
    \null
    \thispagestyle{empty}
    \addtocounter{page}{-1}
    \newpage
    }

\title{Active Learning Enhances Classification of Histopathology Whole Slide Images with Attention-based Multiple Instance Learning }

\name{Ario Sadafi$^{1,2}$ \qquad Nassir Navab$^{2,3}$ \qquad Carsten Marr$^{1}$}

\address{$^{1}$ Institute of AI for Health, Helmholtz Munich, Neuherberg, Germany \\$^{2}$ Computer Aided Medical Procedures (CAMP), Technical University of Munich, Germany\\$^{3}$ Computer Aided Medical Procedures, Johns Hopkins University, USA}

\begin{document}

\onecolumn

\hspace{0pt}
\vfill

{\Huge IEEE Copyright Notice}
\\[0.7in]
{\large © 2023 IEEE. Personal use of this material is permitted. Permission from IEEE must be obtained for all other uses, in any current or future media, including reprinting/ republishing this material for advertising or promotional purposes, creating new collective works, for resale or redistribution to servers or lists, or reuse of any copyrighted component of this work in other works.}
\\[.3in]
\textbf{\large
Pre-print of article that will appear at the 2023 IEEE International Symposium on Biomedical Imaging (ISBI 2023), April 18-21 2023}
\hspace{0pt}
\vfill

\blankpage{}
\twocolumn
\maketitle
\begin{abstract}
In many histopathology tasks, sample classification depends on morphological details in tissue or single cells that are only visible at the highest magnification. For a pathologist, this implies tedious zooming in and out, while for a computational decision support algorithm, it leads to the analysis of a huge number of small image patches per whole slide image (WSI). Attention-based multiple instance learning (MIL), where attention estimation is learned in a weakly supervised manner, has been successfully applied in computational histopathology, but it is challenged by large numbers of irrelevant patches, reducing its accuracy. 
Here, we present an active learning approach to the problem. Querying the expert to annotate regions of interest in a WSI guides the formation of high-attention regions for MIL. We train an attention-based MIL and calculate a confidence metric for every image in the dataset to select the most uncertain WSIs for expert annotation. We test our approach on the CAMELYON17 dataset classifying metastatic lymph node sections in breast cancer. With a novel attention guiding loss, this leads to an accuracy boost of the trained models with few regions annotated for each class.  Active learning thus improves WSIs classification accuracy, leads to faster and more robust convergence, and speeds up the annotation process. It may in the future serve as an important contribution to train MIL models in the clinically relevant context of cancer classification in histopathology.
\end{abstract}
\section{Introduction}
Computational histopathology has emerged as a prototypical clinical application for modern machine learning approaches \cite{coudray2018classification,campanella2019clinical}. The promise is to assist the pathologist in disease classification, and to provide guidance through the gigabyte large scans of a patient's tumor tissue section with a decision support system that alleviates the burden from the ever increasing number of diagnostic tasks \cite{van2021deep}.
In many histopathology diagnosis tasks the necessary information is at single cell and tissue level and requires analysis at high resolution. Since histopathological scans are typically large (on the size of few gigabyte e.g. billions of pixel), this leads to a considerable time investment of the clinical expert for inspecting slides - and a huge number of image patches to be handled by a decision support system. Recently, multiple instance learning (MIL) methods have successfully employed attention mechanisms to focus on certain patches that contain diagnostically relevant information \cite{campanella2019clinical}. MIL is trained in a weakly supervised manner just based on the bag labels of the patients. However, MIL convergence is challenged by the high number of irrelevant patches, especially in cases where the number of training images is limited, like for rare disease or material from pediatric diseases. In some cases it is shown that instance annotations can improve the performance \cite{NIPS2007_a1519de5}.

Our idea is that directing the algorithm with a few region of interest (RoI) annotations could help convergence and enhance model performance and reliability. Obtaining RoI annotations from experts however is tricky, since pathologist's time is scarce and expensive. We thus employ an active learning approach that aims to obtain high performance with fewest possible expert annotations to cut expenses and time. Active learning frameworks estimate the uncertainty of an instance or sample and thus identify most informative data samples. For instance, Sadafi et al. \cite{sadafi2019multiclass} propose an approach for annotation of a Faster R-CNN architecture to detect single red blood cells in brightfield images, improving rare class prediction with Monte Carlo variational inference. Yang et al. \cite{yang2017suggestive} have a statistical approach for estimation of uncertainty by ensemble models and Kim et al. \cite{kim2020confident} propose an active learning method that not only uses uncertainty measurement but also the distribution of the data to select samples for annotation.
Our active learning human-in-the-loop approach selects the most informative WSIs based on a relevance measure that calculates class and pixel uncertainty. Most relevant WSIs are then selected and presented to the expert for RoI annotation. Using these expert annotations for guiding the MIL's attention helps the model to converge. We showcase the usefulness of our approach on the well-known CAMELYON17 dataset \cite{bandi2018detection} by training a completely generic MIL architecture. The task is to detect breast cancer metastasis in lymph node histopathology WSIs. Our work has three main contributions: We are the first to use active learning for RoI annotation on histopathology data, and the first to use the RoI annotations to train an attention-based MIL architecture. Moreover, we can show that attention guiding loss leads to better performance and a more robust model convergence.

\section{Methodology}

\begin{figure}[t]
\centering
\includegraphics[width=0.48\textwidth,page=1,trim=0cm 1.34cm 0cm 0cm,clip]{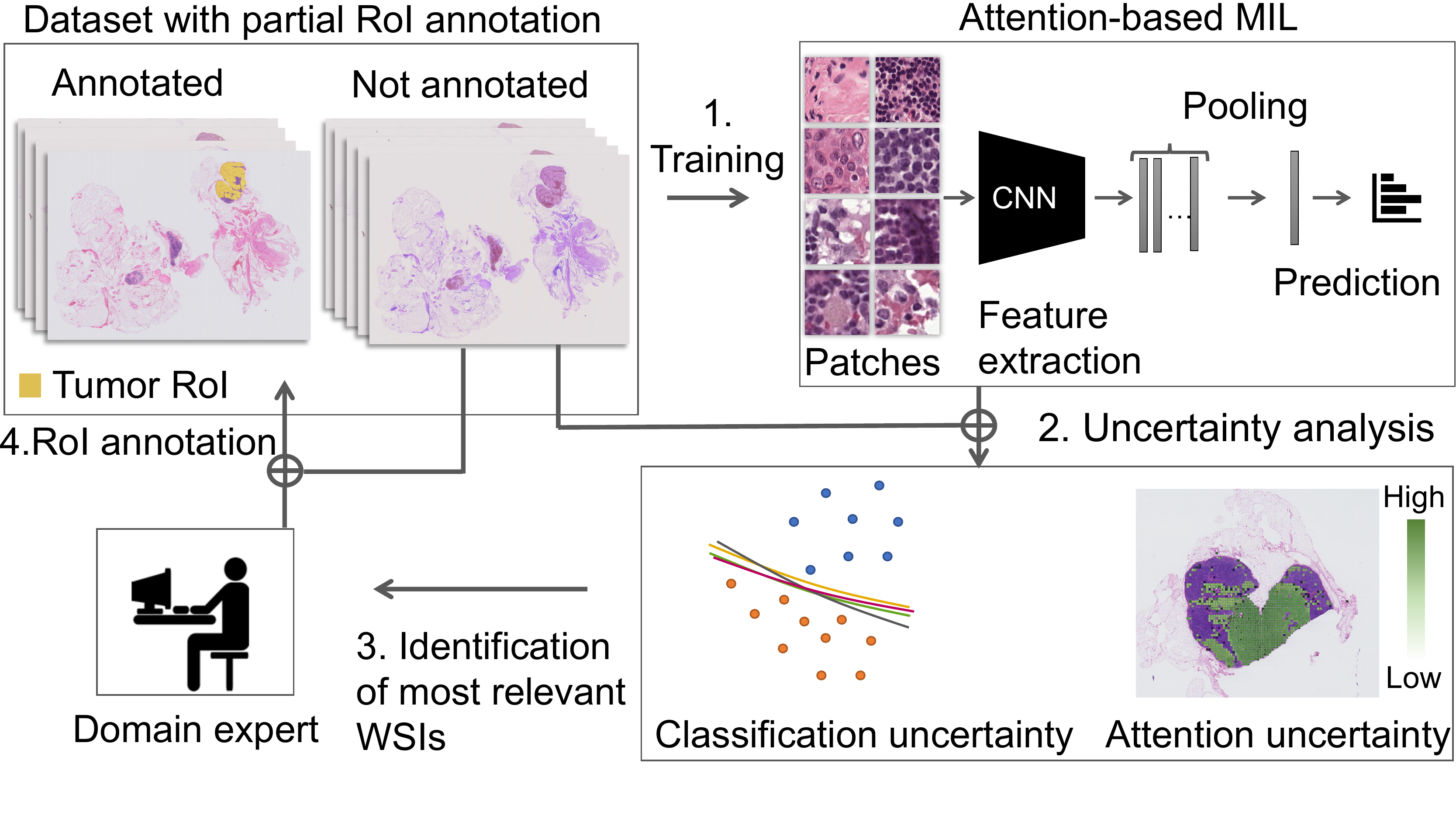}
\caption{Overview of the proposed active learning approach. Our modified multiple instance learning (MIL) with attention guidance is trained on a set of whole slide images (WSIs). In some slides the region of interest (RoI) is annotated by the expert (yellow part). After training, we estimate the attention and classification uncertainty of the model. Next, most relevant WSIs are suggested to the expert for the next round of RoI annotation. }
\label{fig:overview}

\end{figure}

Our proposed active learning human-in-the-loop approach starts with an attention-based MIL \cite{ilse2018attention} model that we train on a dataset of histology whole slide images (WSIs) with bag labels but without any region of interest (RoI) annotation (see Fig \ref{fig:overview}). Using Monte Carlo dropout sampling \cite{gal2016dropout} we then estimate the model’s uncertainty both in final classification and allocation of the attention to image regions  (see Fig \ref{fig:overview}). Based on these two uncertainty scores, unannotated WSIs are sorted and the most relevant ones are passed to the expert for annotation. The expert annotates RoI on the WSI, highlighting tumorous tissue, but not all tumor regions have to be annotated (Fig \ref{fig:overview}). This loop continues by training the model again on the dataset and incorporating the RoI annotation on some of the images provided by the expert. 

In the following, we first explain the classification approach and then our active learning framework. For classification we are adapting the attention-based multiple instance learning (MIL) method proposed by Sadafi et al. \cite{sadafi2020attention} that has an additional single instance classifier branch with an annealing factor.
If $P$ is the set of features of all patches extracted from a WSI, the MIL classifier can be formulated as follows

\begin{equation}
    c_i , \alpha_k = f(P),
\end{equation}
where $c_i \in C$ is the prediction and $\alpha$ is vector of attention values associated with instances. 

\subsection{Preprocessing: Patching and feature extraction}

Patching the WSI and extracting features is the preprocessing step of our method. Any off-the-shelf algorithm can be used for patching and feature extraction. Here, patches of size $512\times512$ are extracted from $40\times$ magnification WSIs. We used Otsu thresholding \cite{otsu1979threshold} in HSV space on saturation to segment the tissue from the background and all patches having less than $50\%$ of tissue coverage are discarded.

For extracting the features of each patch, we used KimiaNet \cite{riasatian2021fine} which is a pretrained DenseNet-121 \cite{huang2017densely} architecture with four dense blocks trained on a huge dataset of histopathological images obtained from TCGA demonstrating a variety of different configurations, tissue types and domains. We extract features by obtaining the output activations of the fourth dense block and perform an adaptive average pooling to reduce dimension and yield feature vectors of size $2048$ for every patch.

\subsection{Attention-based multiple instance learning}
The MIL classifier consists of three main components: (i) feature extraction and embedding of the input instances into a low dimensional representation, (ii) a pooling method to combine all representations, and (iii) a final bag classifier.
Having embedding feature vectors of $h_i = f_\mathrm{EMB}(p_i, \theta): p_i \in P$, MIL approach can be formulated as

\begin{equation}
    \mathcal{L}_\mathrm{MIL} (\theta, \phi) = \mathrm{CE}(c,\hat{c})
\end{equation}
where $c$ is the class associated to the bag, $\hat{c} = \mathrm{f}_{MIL} = (H, \alpha;\phi)$, and $\theta, \phi$ are parameters learned during the training. 
Ilse et al. \cite{ilse2018attention} have introduced an attention pooling method which is a weighted average over all of the instance feature vectors $h_i$ to obtain bag representation $z$

\begin{equation}
    z = \sum_{k=1}^{N} \alpha_k h_k,
\end{equation}
where the weights are estimated by the network as
\begin{equation}
    \alpha_k = \frac{\mathrm{exp}\{w^T\mathrm{tanh}(Vh_k^T)\}}{\sum_{j=1}^N \mathrm{exp}\{w^T\mathrm{tanh}(Vh_j^T)\}}.
\end{equation}
where $V,w \in \phi$ are the learnable parameters. 

One of the known issues with MIL approaches is the vanishing gradients problem \cite{sadafi2020attention,conjeti2017deep} and single instance classification (SIC) is an intuitive method that can help a better convergence such that during the first steps of the training, SIC is the main contributor to the overall loss and as training continues its contribution decreases (see sec \ref{los}). 
We formulate the SIC as
\begin{equation}
    \mathcal{L}_{\mathrm{SIC}}(\theta, \gamma) = \frac{1}{N}\sum_{i=1}^{N}\mathrm{CE}(c_i, \hat{c}_i)
\end{equation}
where $\hat{c}_i = f_{\mathrm{SIC}} (h_i; \gamma)$ is the prediction for instance $i$  and $\gamma$ is the parameters of SIC branch.

\subsection{Attention guiding loss}
Since WSIs are partially annotated 
by the expert, we are proposing a loss that guides the weakly supervised attention mechanism of the MIL to focus on the suggested areas. Attention guiding loss (AGL) is based on a multi-target cross entropy objective on the attention scores regulated after a sigmoid activation function. Formally, if $Q$ is the set of all patches in a RoI, we define attention guiding loss as

\begin{equation}
  \mathcal{L}_{\mathrm{AGL}} (\theta, \phi) = \mathrm{BCE}( \frac{1}{1 + \mathrm{exp}(w^T\mathrm{tanh}(Vh_k^T)}, \mathbf{1}) :h_k \in Q
\end{equation}
where $\mathbf{1}$ is a vector of ones having equal length with $Q$ and BCE is binary cross entropy loss. 
Since the loss is defined only on $Q$, in order to avoid collapse of the attention mechanism, we define $\mathcal{L}_\mathrm{AGL}$ also for the negative cases in dataset $P_{neg}$ as 
\begin{equation}
  \mathcal{L}_{\mathrm{AGL}} (\theta, \phi) = \mathrm{BCE}( \frac{1}{1 + \mathrm{exp}(w^T\mathrm{tanh}(Vh_k^T)}, \epsilon) :h_k \in P_{\mathrm{neg}}
\end{equation}
where $\epsilon$ is a small number close to zero encouraging uniform distribution of the attention on negative WSIs without a tumor. 

\subsection{Overall objective function}\label{los}
The overall objective function of the classification can be defined as 
\begin{equation}
    \mathcal{L}(\theta, \phi, \gamma) = \beta^E\mathcal{L}_{\mathrm{SIC}} + (1-\beta^E)(\mathcal{L}_{\mathrm{MIL}} + \delta\mathcal{L}_{\mathrm{AGL}})
\end{equation}
where $E$ is the epoch number, and $\beta$ and $\delta$ are hyper-parameters regulating the annealing factor for single instance classification and contribution of the attention guiding loss respectively.

\subsection{Uncertainty estimation}
For every unannotated WSI, we measure the epistemic uncertainty of the model through dropout variational inference \cite{kendall2017uncertainties} obtaining two scores: (i) attention uncertainty and (ii) classification uncertainty
both calculated across $N$ inferences.

\subsubsection{Attention uncertainty} 
\begin{figure}[t]
\centering
\includegraphics[width=.48\textwidth,page=2,trim=0cm 2cm 0cm 0cm,clip]{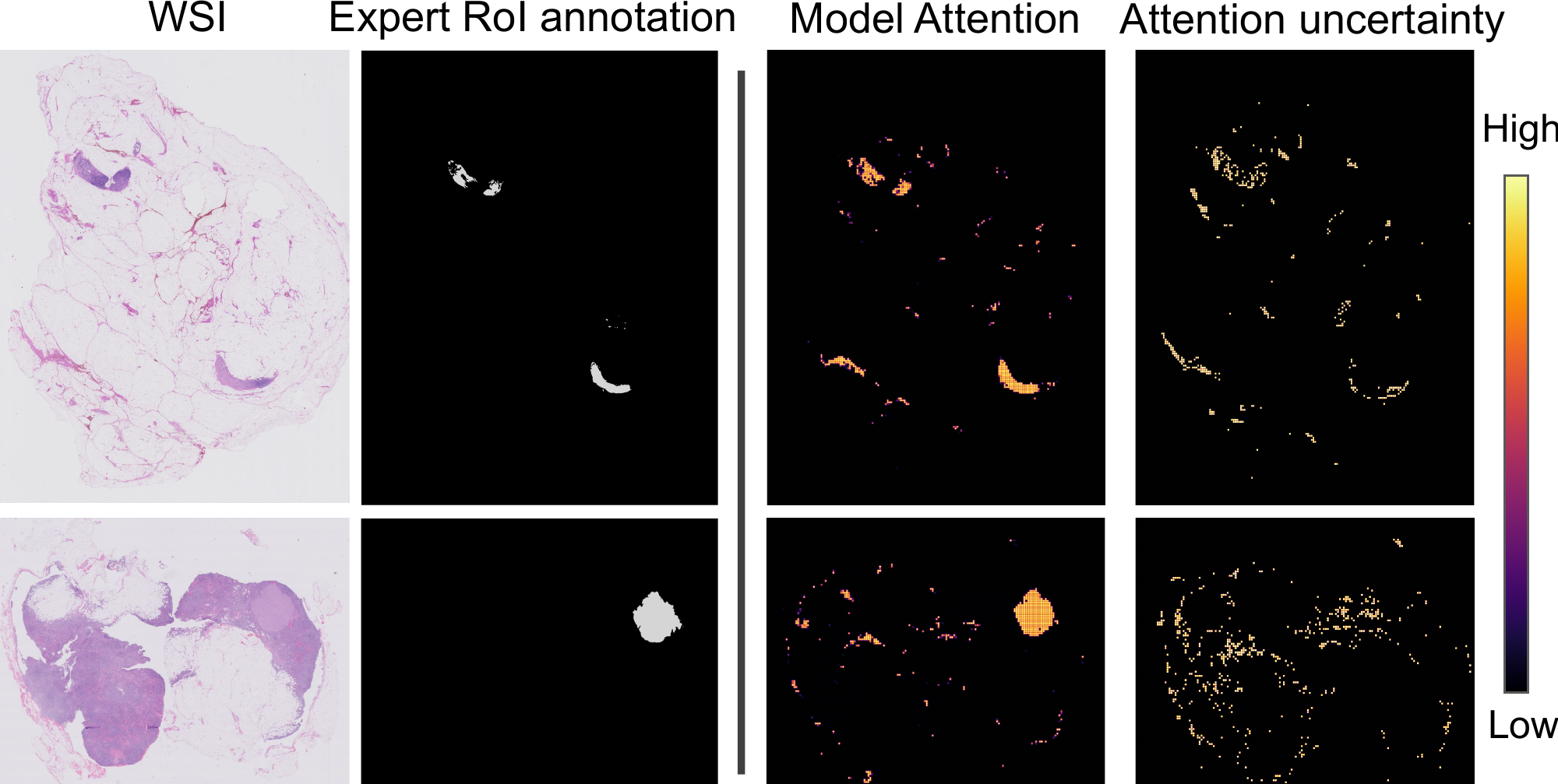}
\caption{Two exemplary WSIs displayed along with expert RoI annotation, attention of the model and estimated attention uncertainty. Model attention is focused but not limited to the expert annotation while attention uncertainty is lower in the annotated regions and higher in the regions which the model finds informative.}
\label{figatt}

\end{figure}

Attention vector is obtained by calculating $\alpha_k = w^T\mathrm{tanh}(Vh_k^T)$. We compare the output values across different inferences and calculate the uncertainty of the WSI based on the mean standard deviation in the attention of every patch as 

\begin{equation}
    U^{att} = \frac{1}{M}\sum_{i=1}^{M}\sqrt{\frac{1}{N}\sum_{k}^{N}(\alpha_{i,k}- \bar{\alpha_{i}})}
\end{equation}
where $M$ is the total number of patches in the image and $\alpha_i$ is the mean attention across $N$ inferences. We then normalize attention uncertainties between 0 and 1. Figure \ref{figatt} shows uncertainty map on a subsection of two exemplary WSIs along with the expert RoI annotation and model's attention. 

\subsubsection{Classification uncertainty}

We pick the most probable class from the softmax output and compare it against the $N$ inferences. Having the groundtruth class $c_{\mathrm{GT}}$, we measure the class uncertainty by
\begin{equation}
    U^{cls} = \frac{1}{N}\sum_{i=1}^{N} [c_i=c_{\mathrm{GT}}]: c_i \in C
\end{equation}
The relevance score for each WSI is calculated based on linear summation of both uncertainties and ranked accordingly.

\section{Experiments \& Results}

\subsection{Dataset}
We are testing the proposed approach on CAMELYON17 dataset \cite{bandi2018detection}. Since currently only the annotations of the training set is publicly accessible, we use that section for our experiments and focus on lesion-level data. Our dataset consists of 100 patients, 500 WSIs, collected from 5 data sources labelled as Macro, Micro, ITC, and negative based on the presence of tumor and its size. Additionally for 50 WSIs lesion annotation of the tumor area (RoI) is provided that we are using it to simulate expert annotation in our experiments.

\subsection{Implementation details}

\textit{Multiple instance learning} architecture consists of a multi-layer neural network made from four linear layers each regulated with leaky ReLU, one dropout layer and batch normalization. The auxiliary single instance classifier and final bag classifier are consisting of 3 and 2 linear layers respectively.

\textit{Training}.
We performed stratified train test split on the available data leaving 34\% of the WSIs for testing the model. Adam optimizer with a learning rate of $5e-5$ is used for training the model for 100 epochs in PyTorch framework. For overall objective calculation $\beta$ is set to $0.7$ and $\delta$ is $0.1$ in all training runs.

\textit{Active learning}. For active learning simulation, at every cycle the expert is asked to annotate two WSIs. Number of samplings ($N$) for uncertainty estimation is 10 times. Each active learning run is continued for 7 queries or 14 WSIs.

\subsection{Attention guiding loss}
One of the main contributions of our design is the attention guiding loss and to study effectiveness of different components of classifier, we have designed an ablation study where we compare our proposed method with other baselines like simple attention-based MIL (MIL) \cite{ilse2018attention}, MIL with auxiliary SIC branch (S-MIL) \cite{sadafi2020attention}, MIL with our attention guiding loss (MIL-AGL) and finally our proposed method of MIL with SIC branch and attention guiding loss (S-MIL-AGL). 
Table \ref{abltbl} shows the results on 10 independent runs over the complete dataset with all 50 RoI annotated WSIs included.

\begin{table}[t]
\caption{Comparison of our proposed method (S-MIL-AGL) with three baselines for WSI classification. Performing 10 runs, mean and standard deviation of accuracy, weighted F1 score and area under ROC is reported.}
\label{abltbl}

\begin{center}
\label{tblablation}
\begin{tabular}{p{2cm}|P{1.6cm}|P{1.6cm}|P{1.6cm}}
\textbf{Method}           & \textbf{Accuracy} & \textbf{F1-Score} & \textbf{AU ROC} \\\hline
MIL         & 0.33 ± 0.27& 0.31 ± 0.30 & 0.54 ± 0.12      \\\hline
S-MIL     & 0.28 ± 0.20& 0.28 ± 0.22 & 0.52 ± 0.12   \\\hline
MIL-AGL     & 0.46 ± 0.24& 0.47 ± 0.23 &  0.59 ± 0.09     \\\hline
S-MIL-AGL & \textbf{0.72 ± 0.03}& \textbf{0.69 ± 0.03} & \textbf{0.64 ± 0.01}     
\end{tabular}
\end{center}

\end{table}

\subsection{Active learning results}
\begin{figure}[t]

\centering
\includegraphics[width=.44\textwidth,page=3,trim=0cm 6.1cm 14cm 0cm,clip]{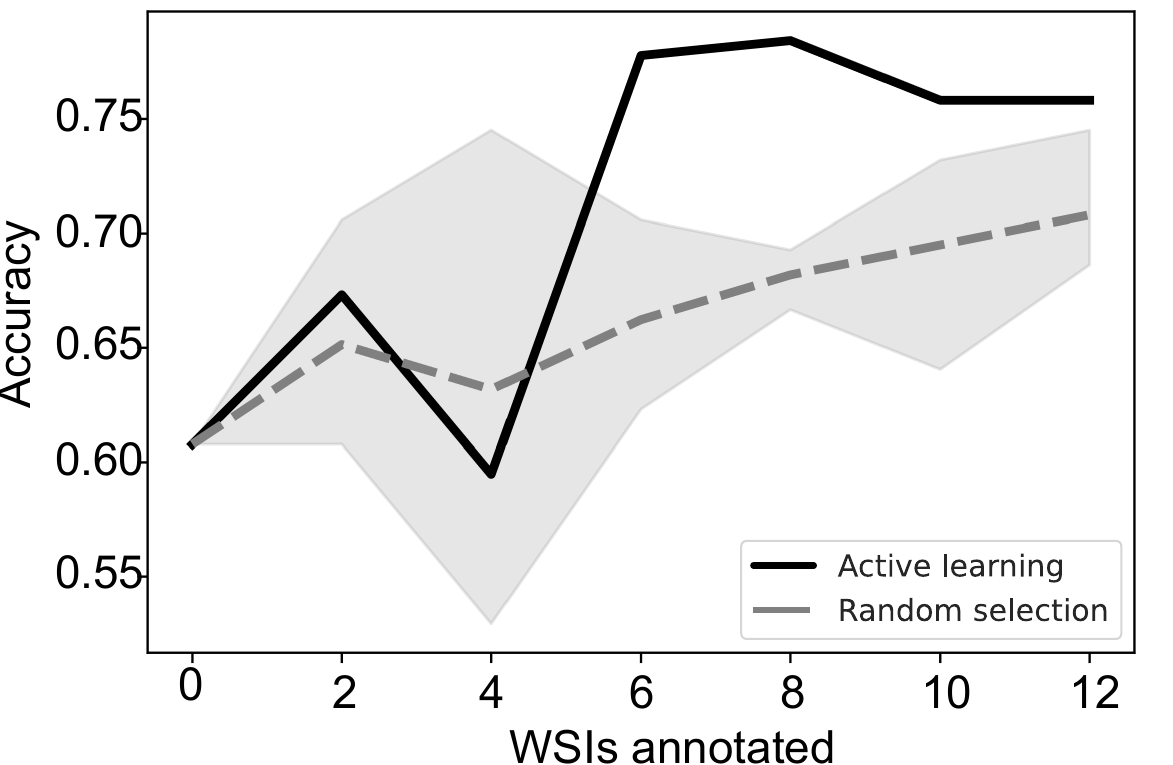}
\caption{Comparison of the model accuracy with number of WSIs annotated. Our active learning approach outperforms random selection of WSIs from CAMELYON17 for annotation and leads to a higher accuracy with fewer number of queries to the expert.}
\label{figAL}
\end{figure}

We simulated the active learning scenario by running the method for 3 complete active learning runs of 7 cycles. At each cycle the expert annotates 2 WSIs and the model is trained not only based on weak bag labels but also using the annotated RoIs. To compare the superiority of our proposed uncertainty based WSI for expert annotation, we are comparing it to when WSIs are selected randomly for annotation. Figure \ref{figAL} shows the accuracy of the model throughout the active learning cycles.

\subsection{Discussion}
Training MIL models for bags containing many instances is not an easy task and can fail. We performed all experiments at $40\times$ magnification, resulting to up to 25,000 patches per sample, most MIL based methods struggle with such a huge number of instances.
Information held within the annotated RoIs is a simple to obtain while highly effective solution in such problems helping the model to identify information rich patches quickly. This way the model converges faster and with lower variance boosting the accuracy of the classification (see Table 1). This robust convergence is also evident in Fig. \ref{figAL}. Additionally, selecting the most informative RoIs is important for fewer annotation by the pathologists whose time is scarce and expensive. We decided to use a completely generic MIL with no task specific tricks or modifications to intensify the important role of our active learning and attention guiding loss in obtaining accurate models faster with better attention allocations and fewer queries from the expert.

\section{Conclusion}
Our novel human-in-the-loop active learning for MIL approach ensures the most informative WSIs with the highest impact on the training are annotated by the experts. WSIs are not regularly annotated at the lesion level and this approach will help create datasets with the least effort and cost. Our novel attention guiding loss also helps convergence of the attention module on these informative patches boosting the performance of the MIL classifier by helping it to discover the most informative patches. Development of software-based systems supporting the active learning process and improving other similar more recent classification approaches such as transformer-based methods to train with RoI annotation are some of the possible future works.

\subsection*{Compliance with Ethical Standards}
This research study was conducted retrospectively using human subject data made available in open access by CAMELYON17 \cite{bandi2018detection}. Ethical approval was not required as confirmed by the license attached with the open access data.
\subsection*{Acknowledgements}
C.M. has received funding from the European Research Council (ERC) under the European Union’s Horizon 2020 research and innovation programme (Grant agreement No. 866411)
\bibliographystyle{IEEEbib}
\bibliography{strings,refs}

\begin{thebibliography}{10}

\bibitem{coudray2018classification}
Nicolas Coudray, Paolo~Santiago Ocampo, Theodore Sakellaropoulos, Navneet
  Narula, Matija Snuderl, David Feny{\"o}, Andre~L Moreira, Narges Razavian,
  and Aristotelis Tsirigos,
\newblock ``Classification and mutation prediction from non--small cell lung
  cancer histopathology images using deep learning,''
\newblock {\em Nature medicine}, vol. 24, no. 10, pp. 1559--1567, 2018.

\bibitem{campanella2019clinical}
Gabriele Campanella, Matthew~G Hanna, Luke Geneslaw, Allen Miraflor, Vitor
  Werneck Krauss~Silva, Klaus~J Busam, Edi Brogi, Victor~E Reuter, David~S
  Klimstra, and Thomas~J Fuchs,
\newblock ``Clinical-grade computational pathology using weakly supervised deep
  learning on whole slide images,''
\newblock {\em Nature medicine}, vol. 25, no. 8, pp. 1301--1309, 2019.

\bibitem{van2021deep}
Jeroen Van~der Laak, Geert Litjens, and Francesco Ciompi,
\newblock ``Deep learning in histopathology: the path to the clinic,''
\newblock {\em Nature medicine}, vol. 27, no. 5, pp. 775--784, 2021.

\bibitem{NIPS2007_a1519de5}
Burr Settles, Mark Craven, and Soumya Ray,
\newblock ``Multiple-instance active learning,''
\newblock in {\em Advances in Neural Information Processing Systems}, J.~Platt,
  D.~Koller, Y.~Singer, and S.~Roweis, Eds. 2007, vol.~20, Curran Associates,
  Inc.

\bibitem{sadafi2019multiclass}
Ario Sadafi, Niklas Koehler, Asya Makhro, Anna Bogdanova, Nassir Navab, Carsten
  Marr, and Tingying Peng,
\newblock ``Multiclass deep active learning for detecting red blood cell
  subtypes in brightfield microscopy,''
\newblock in {\em International Conference on Medical Image Computing and
  Computer-Assisted Intervention}. Springer, 2019, pp. 685--693.

\bibitem{yang2017suggestive}
Lin Yang, Yizhe Zhang, Jianxu Chen, Siyuan Zhang, and Danny~Z Chen,
\newblock ``Suggestive annotation: A deep active learning framework for
  biomedical image segmentation,''
\newblock in {\em International conference on medical image computing and
  computer-assisted intervention}. Springer, 2017, pp. 399--407.

\bibitem{kim2020confident}
Seong~Tae Kim, Farrukh Mushtaq, and Nassir Navab,
\newblock ``Confident coreset for active learning in medical image analysis,''
\newblock {\em arXiv preprint arXiv:2004.02200}, 2020.

\bibitem{bandi2018detection}
Peter Bandi, Oscar Geessink, Quirine Manson, Marcory Van~Dijk, Maschenka
  Balkenhol, Meyke Hermsen, Babak~Ehteshami Bejnordi, Byungjae Lee, Kyunghyun
  Paeng, Aoxiao Zhong, et~al.,
\newblock ``From detection of individual metastases to classification of lymph
  node status at the patient level: the camelyon17 challenge,''
\newblock {\em IEEE transactions on medical imaging}, vol. 38, no. 2, pp.
  550--560, 2018.

\bibitem{ilse2018attention}
Maximilian Ilse, Jakub Tomczak, and Max Welling,
\newblock ``Attention-based deep multiple instance learning,''
\newblock in {\em International conference on machine learning}. PMLR, 2018,
  pp. 2127--2136.

\bibitem{gal2016dropout}
Yarin Gal and Zoubin Ghahramani,
\newblock ``Dropout as a bayesian approximation: Representing model uncertainty
  in deep learning,''
\newblock in {\em international conference on machine learning}. PMLR, 2016,
  pp. 1050--1059.

\bibitem{sadafi2020attention}
Ario Sadafi, Asya Makhro, Anna Bogdanova, Nassir Navab, Tingying Peng, Shadi
  Albarqouni, and Carsten Marr,
\newblock ``Attention based multiple instance learning for classification of
  blood cell disorders,''
\newblock in {\em International Conference on Medical Image Computing and
  Computer-Assisted Intervention}. Springer, 2020, pp. 246--256.

\bibitem{otsu1979threshold}
Nobuyuki Otsu,
\newblock ``A threshold selection method from gray-level histograms,''
\newblock {\em IEEE transactions on systems, man, and cybernetics}, vol. 9, no.
  1, pp. 62--66, 1979.

\bibitem{riasatian2021fine}
Abtin Riasatian, Morteza Babaie, Danial Maleki, Shivam Kalra, Mojtaba Valipour,
  Sobhan Hemati, Manit Zaveri, et~al.,
\newblock ``Fine-tuning and training of densenet for histopathology image
  representation using tcga diagnostic slides,''
\newblock {\em Medical Image Analysis}, vol. 70, pp. 102032, 2021.

\bibitem{huang2017densely}
Gao Huang, Zhuang Liu, Laurens Van Der~Maaten, and Kilian~Q Weinberger,
\newblock ``Densely connected convolutional networks,''
\newblock in {\em Proceedings of the IEEE conference on computer vision and
  pattern recognition}, 2017, pp. 4700--4708.

\bibitem{conjeti2017deep}
Sailesh Conjeti, Magdalini Paschali, Amin Katouzian, and Nassir Navab,
\newblock ``Deep multiple instance hashing for scalable medical image
  retrieval,''
\newblock in {\em International Conference on Medical Image Computing and
  Computer-Assisted Intervention}. Springer, 2017, pp. 550--558.

\bibitem{kendall2017uncertainties}
Alex Kendall and Yarin Gal,
\newblock ``What uncertainties do we need in bayesian deep learning for
  computer vision?,''
\newblock {\em Advances in neural information processing systems}, vol. 30,
  2017.

\end{thebibliography}

\end{document}